\documentclass[letterpaper]{article} 
\usepackage{aaai24}  
\usepackage{times}  
\usepackage{helvet}  
\usepackage{courier}  
\usepackage[hyphens]{url}  
\usepackage{graphicx} 
\usepackage{natbib}  
\usepackage{caption} 
\usepackage{algorithm}
\usepackage{array} 
\usepackage[table,xcdraw]{xcolor}
\usepackage{algorithmicx}
\usepackage{algpseudocode}
\usepackage{color,xcolor}
\usepackage{color}
\usepackage{amssymb}
\usepackage{bbding}
\usepackage{threeparttable}
\usepackage{booktabs}
\usepackage{multirow}
\usepackage{graphicx}
\usepackage{subfigure}
\usepackage{multicol}
\usepackage{lipsum}
\usepackage{multirow}
\usepackage{graphicx}
\usepackage[table,xcdraw]{xcolor}

\usepackage{amsmath}
\usepackage{amssymb}
\usepackage{mathtools}
\usepackage{amsthm}
\usepackage{multicol,mwe,float}

\usepackage{xargs}                      
\newcommand{\zj}[1]{\textcolor{black}{#1}}

\usepackage{newfloat}
\usepackage{listings}
\DeclareCaptionStyle{ruled}{labelfont=normalfont,labelsep=colon,strut=off} 
\floatstyle{ruled}
\newfloat{listing}{tb}{lst}{}
\floatname{listing}{Listing}
\title{CasIL: Cognizing and Imitating Skills via a Dual Cognition-Action Architecture}
\author{Zixuan Chen$^{1}$, Ze Ji$^{2}$, Shuyang Liu$^{1}$, Jing Huo$^{1}$, Yiyu Chen$^1$, Yang Gao$^1$\\
{\normalsize{$^1$State Key Laboratory for Novel Software Technology, Nanjing University, Nanjing, China}}\\
{\normalsize{$^2$School of Engineering, Cardiff University, Cardiff, U.K.}}}

\begin{document}

\maketitle

\begin{abstract}
Enabling robots to effectively imitate expert skills in long-horizon tasks such as locomotion, manipulation, and more, poses a long-standing challenge. Existing imitation learning (IL) approaches for robots still grapple with sub-optimal performance in complex tasks. In this paper, we consider how this challenge can be addressed within the human cognitive priors. Heuristically, we extend the usual notion of action to a dual \textit{Cognition (high-level)-Action (low-level)} architecture by introducing intuitive human cognitive priors, and propose a novel skill IL architecture through human-robot interaction, called \textbf{C}ognition-\textbf{A}ction-based \textbf{S}kill \textbf{I}mitation \textbf{L}earning (\textbf{CasIL}), for the robotic agent to effectively cognize and imitate the critical skills from raw visual demonstrations. 
CasIL enables both cognition and action imitation, while high-level skill cognition explicitly guides low-level primitive actions, providing robustness and reliability to the entire skill IL process.
We evaluated our method on MuJoCo and RLBench benchmarks, as well as on the obstacle avoidance and point-goal navigation tasks for quadrupedal robot locomotion. Experimental results show that our CasIL consistently achieves competitive and robust skill imitation capability compared to other counterparts in a variety of long-horizon robotic tasks.
\end{abstract}

\section{Introduction}
In recent years, there is a growing anticipation for intelligent robots to master a broad spectrum of complex skills needed for executing long-horizon tasks \cite{peters2012robot,odesanmi2023skill,li2021human,hiranaka2023primitive}. The ability to learn these skills from expert demonstrations, mimicking the human learning process, is an increasingly desired capability. These skills can range from a humanoid robot maintaining balance, a robotic arm performing dexterous manipulation tasks, to quadrupedal robot  robots achieving point-to-point stable locomotion.
Imitation learning (IL) has proven to be a significant tool in robot learning, allowing agents to learn diverse skill policies based on expert demonstrations \cite{williams1989learning,atkeson1997robot,ho2016generative,fang2019survey,hua2021learning,celemin2022interactive}.

IL in robotics is commonly based on image or video-based expert demonstrations, where visual sensors (e.g., cameras) are used to record the demonstrator's movement, deep learning models then convert these captured movements into control signals for the robot  \cite{finn2017one,du2022play,young2021visual}. This promises robots a variety of skills including, but not limited to, basic locomotion, manipulation, and dynamic obstacle avoidance.
However, achieving such highly intelligent skill imitation often requires a large number of expert demonstrations that are already annotated with high-quality segmented language instructions, and such regulated expert demonstrations tend to be labour-intensive and can be cumbersome to organise and provide in complex tasks \cite{fang2019survey,jing2021adversarial}. Instead, the goal of our work is how to enable robots to learn skills by imitating unlabeled visual demonstrations without any language segmentation annotations, thus achieving higher robustness and availability in a variety of long-horizon  robotic tasks.

To this end, our attention is focused on the hierarchical imitation learning (HIL) architecture.
HIL is gaining increasing recognition as an effective approach to address the difficulties encountered in traditional IL demonstration pre-processing. Utilizing a layered architecture, HIL facilitates the robotic agent to obtain two policy levels from the raw expert demonstrations \cite{jing2021adversarial,zhang2021explainable}. At \zj{the} lower level, robots acquire specialized skills, or sub-policies, to perform specific parts of complex tasks. At \zj{the} high level, robots are able to discern policies that help them switch between different skills in response to the current task environment in order to solve the overall long-horizon task. 
However, a major obstacle to such hierarchical-based learning approaches is that effective learning relies on the robustness and rationality of the hierarchical structure: poor hierarchical logic leads to poor imitation.
This situation is even more severe in robotic applications that rely on visual inputs \cite{peng2018sfv}- if the visual demonstration is low-data or noisy then the hierarchical architecture of the agent's self-exploratory generation can be severely impaired - thus limiting the reliability and applicability of such approaches.

\begin{figure}[htb!]
\centering
\includegraphics[width=0.85\columnwidth]{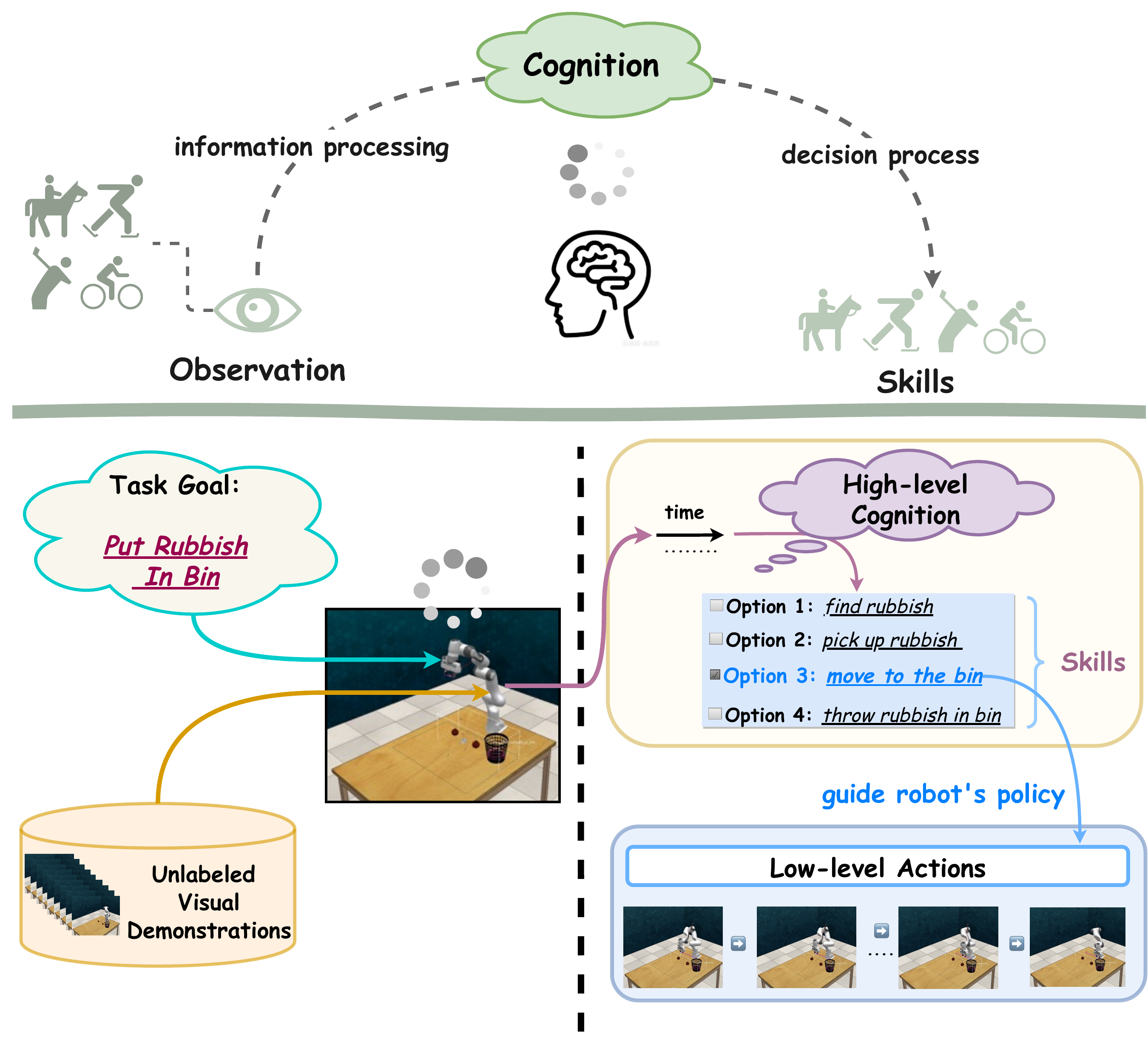} 
\caption{\textbf{Overview of CasIL}. Take the \textit{PutRubbishInBin} task on RLBench benchmark as an illustrative example. Given a task goal and unlabeled visual demonstrations, the robotic agent follows human cognitive priors, utilizes CasIL to generate a task cognitive option chain, and selects the current task cognitive skill \textit{``move to the bin''} from the option chain. A skill-conditioned policy then executes the corresponding actions to perform the skill. }
\label{overview}
\end{figure}
Note the limitations of traditional HIL methods relying only on deep network models for self-exploration to build hierarchy, we take a fresh look at the problem from the perspective of human-robot interaction, introduce human cognitive priors of task decomposition, and exploit the decision-making process of the robot in expert visual demonstrations, with the ambition of developing a more reasonable HIL architecture that significantly enhances the robot's skill IL capability and robustness under long-horizon tasks.
To achieve this, we propose \textbf{C}ognition-\textbf{A}ction-based \textbf{S}kill \textbf{I}mitation \textbf{L}earning (\textbf{CasIL}). CasIL facilitates the learning of explicit cognition and critical skills from offline visual expert demonstrations.
As depicted in Fig.~\ref{overview}, CasIL empowers a robotic arm to generate cognition from unlabeled visual demonstrations. Uniquely, CasIL draws inspiration from the human cognition process for skill imitation and implements a dual cognition-action architecture. The high-level cognition generator aligns textual forms of human cognition of task decomposition with expert visual demonstrations, allowing the robot to create a cognitive option chain, mimicking human cognitive processes to predict and switch between critical skills as subtasks evolve. Meanwhile, the low-level component predicts and executes potential encoded primitive actions inherent in these skills.
Intuitively, high-level cognition acts as a ``manager" that abstracts logical critical skills and provides ``guidance" for the robotic agent to switch between skills. Notably, as a result of processing aligned with human cognition, these critical skills are reliable and can be combined to robustly handle long-horizon tasks.
We evaluate our skill imitation architecture using MuJoco \cite{todorov2012mujoco} and RLBench \cite{james2020rlbench} benchmarks, achieving leading-edge performance in both humanoid robot environments and a variety of challenging robotic arm manipulation environment settings. In addition, we conducted simulation experiments using Bullet Physics \cite{coumans2015bullet} for obstacle avoidance and point-goal navigation tasks with quadrupedal robot locomotion to evaluate the imitation performance of the method in dynamic environments. The results show that CasIL is able to deliver more competitive results compared to other counterparts.

Specifically, Our key contributions include: (1) We innovatively combine human cognitive priors with robotic skill IL processes to propose CasIL, a dual cognition-action IL architecture through human-robot interaction, which enhances robots' skill IL capability in complex long-horizon tasks by learning critical skills with text-image-aligned cognition.
(2) We validate the effectiveness of CasIL on a variety of robotic long-horizon control tasks, including humanoid robots, robotic arms, and quadrupedal robots, demonstrating that by aligning with human cognition of decision-making, robotic agents equipped with the CasIL architecture are significantly more capable of imitating skills from expert demonstrations and are robust to robotic tasks with different morphologies or in low-expert-data regime.


\section{Related Work}
This section begins with an introduction to the classic IL methods for robots\zj{, followed by discussions on related HIL methods and skill-based mechanisms, which provide inspiration for our proposed approach}. 

\subsection{Imitation Learning for Robotics}
Using the imitation learning (IL) mechanism, the robot learns \zj{operation} 
skills by observing an expert's demonstration and generalizes the skills to other unseen scenarios. This process not only extracts information about the behavior and the surrounding environment, but also learns the mapping relationship between observation and performance \cite{hua2021learning}. Under the assumption of IL, the robotic control task can be regarded as a Markov decision process (MDP) \cite{sutton2018reinforcement}, and then the expert's action sequences are encoded into state-action pairs that are consistent with the expert, and IL allows robots to be trained based on previously observed experiences rather than learning from scratch, so the learning efficiency is further improved \cite{bojarski2016end}. In the complete process of robot IL, expert demonstration\zj{s} provide
 teaching sample\zj{s} that contain 
rich features, and skill representations characterize the features in the teaching sample into a valid form that the robot can recognize. The ultimate goal of IL is for the robot to ``master'' the behavior, i.e., the robot has to reproduce the behavior and generalize it to new scenarios. The process by which a robot ``masters'' behavior using schematic information can be referred to as skill imitation \cite{fang2019survey}. Currently, IL for robotics can be similarly categorized into behavioral cloning (BC) \cite{pomerleau1988alvinn,atkeson1997robot}, inverse reinforcement learning (IRL) and generative adversarial imitation learning (GAIL) \cite{ho2016generative,kumar2016learning}.
However, the amount of expert demonstration data that can be collected in many complex long-horizon robot control tasks tends to be in a low-data regime, which prevents traditional ILs from consistently maintaining good imitation performance as the task duration increments.

\subsection{Hierarchical Imitation Learning}
Considering the way in which long-horizon activities are broken down into sub-tasks, HIL can perform better than IL by first developing sub-policies for completing the precise control of each sub-task and then mastering sub-policies for scheduling sub-policies. The options architecture suggested in \cite{sutton1999between} can be used to model the sub-policies in RL. The architecture broadens the definition of action \zj{space} to include options (i.e., closed-loop policies for performing actions over a period of time).
Based on the above task decomposition mechanism, the current mainstream HIL research can be classified into two main categories, namely, hierarchical behavioral cloning (HBC)-type methods by developing sub-policies and hierarchical inverse reinforcement learning (HIRL)-type methods by fusing IL with options \cite{jing2021adversarial}.
Using supervised learning of the appropriate state-action pairs, HBC trains a policy for each sub-task, necessitating the use of expert demonstrations to provide sub-task annotations \cite{le2018hierarchical,kipf2019compile,zhang2021provable}. In HIRL, the option occupancy measurement replaces the occupancy measurement in GAIL, which measures the distribution of state-action pairs. This is to encourage hierarchical policies to generate state-action-option distributions that are similar to the expert demonstration distributions \cite{jing2021adversarial,henderson2018optiongan}. However, the nature of the one-step option-based assumption leads HIRL's hierarchical policy to judge the matching option of the moment at each time step, which significantly increases the computational cost of the algorithm under complex visual input tasks, and its performance depends heavily on the distribution of options inferred by self-exploration \cite{jing2021adversarial}. 
Overall, current HIL approaches either require large amounts of expert data to provide the robot with learning experiences or rely solely on deep network models for self-exploration, limitations that make it difficult for robots to achieve human-like IL paradigm in complex long-horizon tasks. This motivates us to consider whether we can capitalize on the different strengths of these two broad classes of methods and re-explore a mechanism for robot skill imitation with human-like decision logic.

\subsection{Skill-based Mechanisms}
Skill-based mechanisms are widely used in reinforcement learning (RL), and the concept of skills provides valuable insights into the generalization of algorithms. Specifically, skill-based RL generally utilizes task-agnostic experience to accelerate the learning process \cite{hausman2018learning,lee2019learning,merel2018neural}.To extract skills from task-agnostic datasets, there are a number of ways to do this by learning the cognitive embedding of skills and skill priors from existing datasets \cite{pertsch2021accelerating,pertsch2022guided}. 
Inspired by this skill-learning mechanism and in order to propose a hierarchical architecture that is more in alignment with human decision logic, we directly adopt human cognition as a skill prior and apply the human cognitive prior to facilitate the robot's cognition and imitation of the key skills demonstrated by the visual expert, which in a sense realizes the human-robot interaction process.

\section{Our Method: CasIL}
In this section, we first outline the overview of our approach. Subsequently, we describe in detail our \textbf{C}ognition-\textbf{A}ction-based \textbf{S}kill \textbf{I}mitation \textbf{L}earning (\textbf{CasIL}), including the task problem formulation, the method of how to build an option chain embedded with the critical skills through the cognition generator, and the training architecture of how to integrate the generated cognitive skills into the IL process.

\subsection{Overview}

In order to realize the robot's dual imitation \zj{capability} in the cognition-action space, we achieve the integration of cognition and action by introducing \textbf{options} \cite{sutton1999between}. Specifically, CasIL aligns human cognition in textual form with expert visual demonstrations and models the problem as a continuous cognition-action space to line up with the demonstrated actions, 
to more closely approximate human-like decision logic. The key idea of CasIL is to learn text-image-aligned skill embeddings that contain both human cognitive prioris and expert behavioral information, which provides stable and reliable decision-making experiences for robotic skill imitation.  The architecture of our proposed method is illustrated in Fig.~\ref{archi}.

\subsection{Cognition-Action\zj{-}based Skill Imitation Learning}
\begin{figure}[htb]
\centering
\includegraphics[width=0.7\columnwidth]{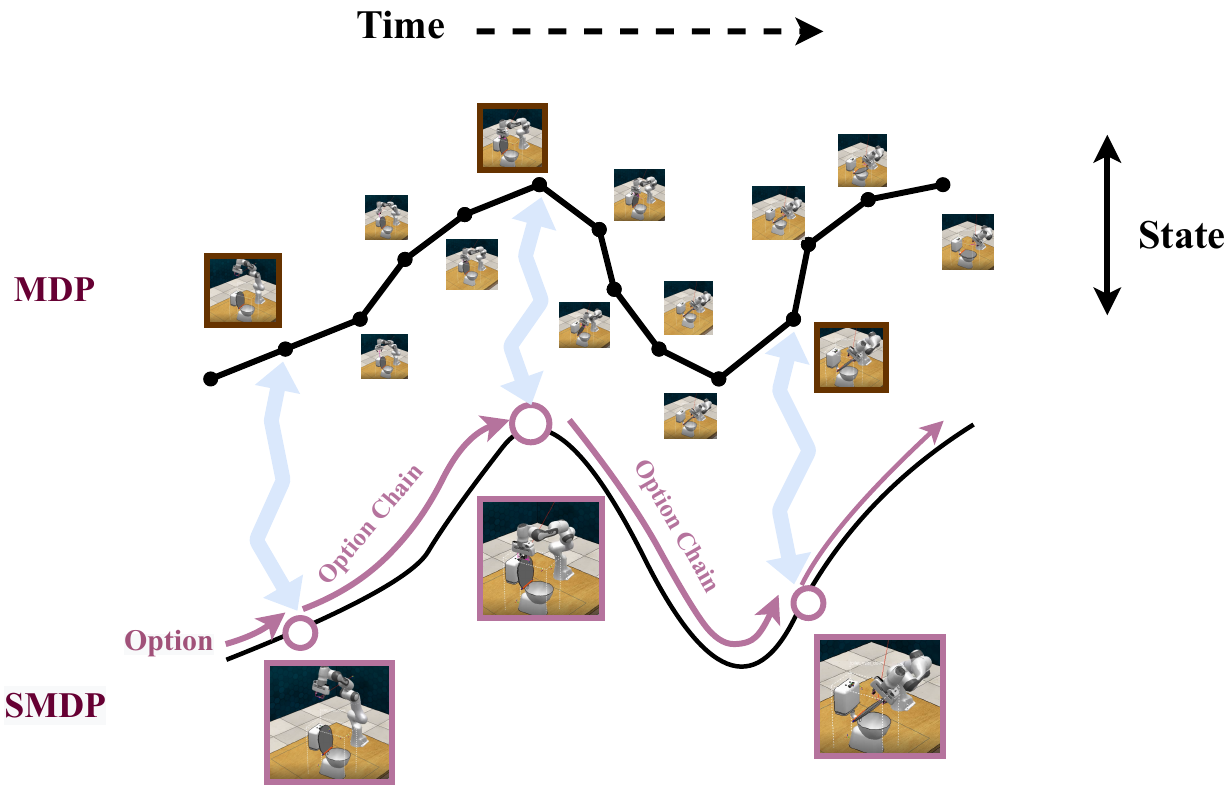} 
\caption{\textbf{Visualization of an SMDP in a robotic arm manipulation task}. Options enable an SMDP trajectory to be temporally extended.}
\label{setup}
\end{figure}

\paragraph{Problem Formulation}
We consider general long-horizon task environments, represented as a Semi-Markov Decision Process (SMDP)\footnote{Please refer to the Appendix for more specific preliminaries.}, which is defined as a tuple $(\mathcal{S},\mathcal{A},{\{\mathcal{I}_o,\mathcal{\pi}_o,\mathcal{\beta}_o\}_{o \in \mathcal{O},}},\pi_{\mathcal{O}(o|s)},\mathcal{P},\mathcal{R})$, where, $\mathcal{S},\mathcal{A},\mathcal{P},\mathcal{R}$ are defined 
\zj{in} the same \zj{manner} as MDP. $\{\mathcal{I}_o,\mathcal{\pi}_o,\mathcal{\beta}_o\}$ denotes an option, which consists of a policy $\pi_o : \mathcal{S} \times \mathcal{A} \rightarrow [0,1]$, a termination condition $\mathcal{\beta}_o: \mathcal{S}^{+} \rightarrow [0,1]$, and an initiation set $ \mathcal{I}_o \in \mathcal{S}$, and an option is available in state $s_t$ if and only if $s_t \in \mathcal{I}$. When the previous option terminates stochastically according to $\mathcal{\beta}$, a new option will be activated by the inter-option policy $\pi_{\mathcal{O}(o|s)}$ until the entire long-horizon task is completed, constituting the option chain.
As shown in Fig.{~\ref{setup}}, SMDP is a special kind of MDP suitable for modelling continuous-time discrete-event systems, where the action execution time within each option is not fixed and can be flexibly varied according to the dynamic characteristics of different tasks, which 
\zj{aligns} with our dual cognition-action imitation architecture.

We assume that a task $\mathcal{T}$ consists of multiple options, each encoding a skill. For example, in the \textit{ToiletSeatDown} task shown in Fig.{~\ref{setup}}, intuitively, this long-horizon task consists of three sub-tasks, each symbolising a skill -- \textit{``grasp the toilet lid''}, \textit{``move the toilet lid downward''}, \textit{``release the toilet lid''} -- in a rigorous logical hierarchical order. 
We assume that we have access to an offline expert dataset $\mathbb{D}$ of trajectories obtained from optimal policies for various long-horizon tasks. Each trajectory $\tau_i = ({G,(s_{i_1}, a_{i_1}), (s_{i_2}, a_{i_2}), \zj{\cdots}, (s_{i_T}, a_{i_T})})$ consists of the final goal $G$ of the task $\mathcal{T}$, 
the observations $s_{i_t} \in \mathcal{S}$ (we will use $s \in \mathcal{S}$ to refer to the observation to be closely related to the task), and actions $a_{i_t} \in \mathcal{A}$ 
\zj{of} $T$ time steps. The trajectories are not labelled with any rewards or language instructions.

Thus, under the problem formulation of our work, each trajectory in the training dataset across different tasks can be combined according to the best skill set (option chian) given human cognitive priors. For example, if the training data is the expert trajectory of \textit{PutRubbishInBin} task, then in terms of human cognitive intuition, the optimal skill set combination of how the task should be done is: \textit{find rubbish}, then \textit{pick up rubbish}, and \textit{move to the bin}, finally \textit{throw rubbish in bin}. Our goal is for the robot to imitate and generate such an optimal skill set, i.e., a option chain in CasIL, and to be trained by CasIL to allow the robotic arm to perform the new \textit{PutRubbishInBin} task well during testing. It is worth noting that the training dataset does not provide the robot with annotated information about where one skill ends and another begins. CasIL must enable the robot to learn how to cognize and organize these skills learned in training, guided by human prioris (i.e., intuitively how the task should be done), and thus achieve a dual cognition-action imitation capability shown in Fig.{\ref{overview}} at test time.  
\begin{figure}[htb!]
\centering
\includegraphics[width=0.8\columnwidth]{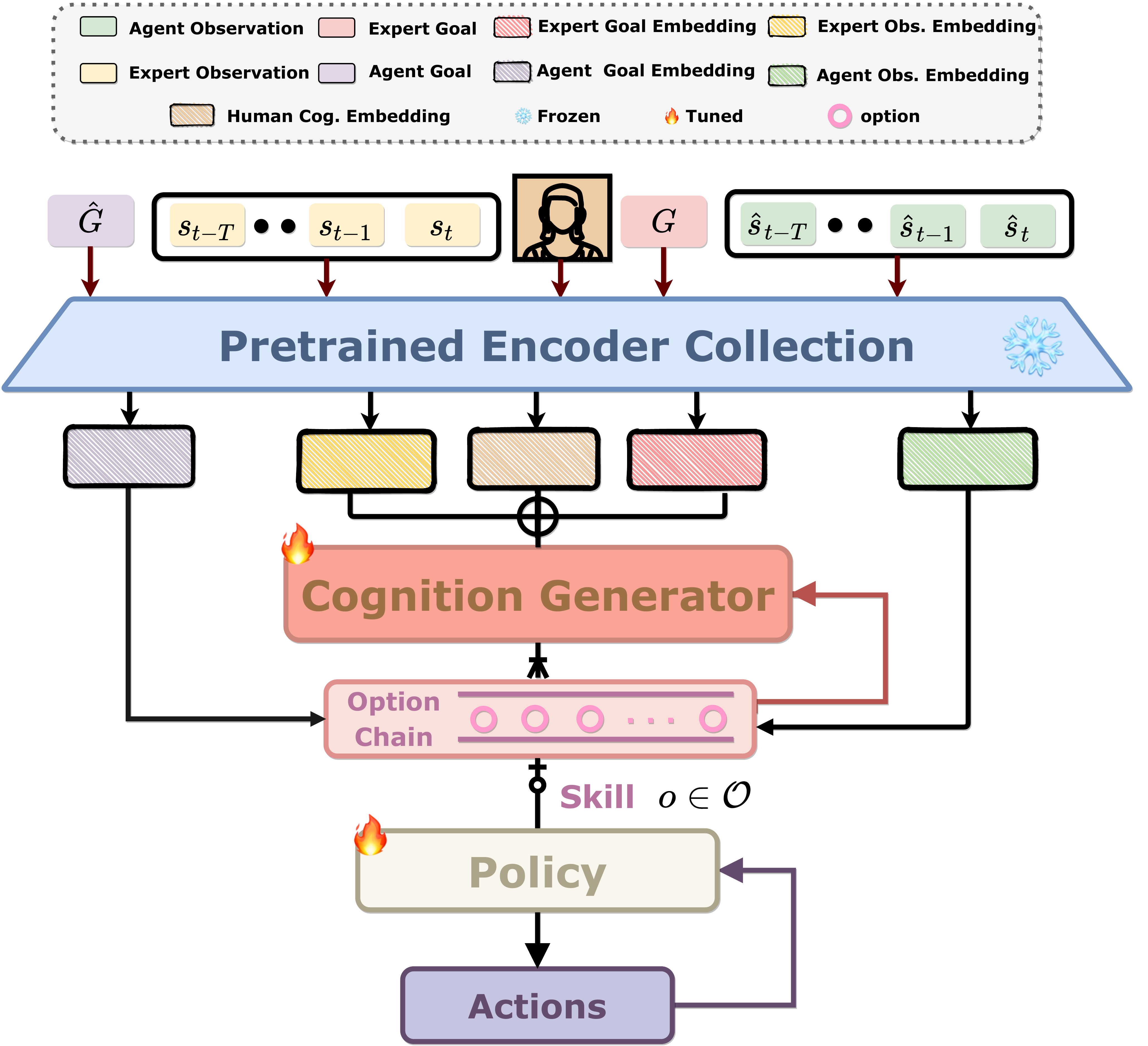} 
\caption{\textbf{CasIL Architecture.} After processing through the pretained image and text encoders, the expert task goal, textual descriptions of human cognitive prioris, and a set of observations are fed into the cognition generator. The generated option chain selects appropriate skills to encode the current sub-task options based on the agent's observation history and the agent's task goals, and feeds them into the policy layer to guide the robotic agent towards a set of action instances. CasIL is trained end-to-end.}
\label{archi}
\end{figure}
  
\paragraph{Skill Abstractions with Generated Cognition}
CasIL's workflow is visualized in Fig.{~\ref{archi}}. Our architecture consists of three working modules from top to bottom: A collection of image and text encoders that encode image observations and text information, respectively. A cognition generator $\mathcal{F} : G \times \mathcal{S} \to \mathcal{O}$ and a policy module $\pi_{\mathcal{O}} : \mathcal{S} \times \mathcal{O} \to \mathcal{A}$. Here, $\mathcal{O} = \{o^{\textbf{1}}, \dots, o^{\textbf{K}} \}$ is a learnable option chain consisting of $\textbf{K}$ options embedded with latent skill. 
The main idea of the above working modules is to divide the robot's skill-based IL into two stages: 1) Learning from expert visual demonstrations and expert task goals that represent the option chain $\mathcal{O}$ of various types of skills in a long-horizon task, and decomposing the task into smaller sub-goals that fit into the logic of human thinking; and 2) conditioning on these options $\mathcal{O} = \{o^{\textbf{1}}, \dots, o^{\textbf{K}} \}$ based on the robotic agent's observations and the task goal that the agent receives to learn policies $\pi_{\mathcal{O}}$. The above stages are end-to-end training, and the details of each module are presented in the Appendix.

Given the expert demonstration input $\tau^E = (G, \{s_t, a_t\}_{t=1}^T)$, and the textual descriptions of observations under critical decisions in the logic of human cognition $\{l_{\textbf{1}},\dots,l_{\textbf{K}}\}$ (e.g., in the \textit{PutRubbishInBin} task, the textual description of the observations, $l_\textbf{1}$, is ``\textit{robotic arm determines location of rubbish and finds rubbish}''). If $s_t$ and $l_\textbf{t}$ are aligned, the cognition generator $\mathcal{F}$ generates critical cognitive embeddings of the decision logic in the expert demonstration, i.e., option $o^{\textbf{t}} \in G \times \mathcal{S},  \textbf{1} \leqslant \textbf{t} \leqslant \textbf{K}$ at each optimal skill division time step $\bar{t}$, i.e., $o^{\textbf{t}} = \mathcal{F}(G, (s_{\bar{t}}, \dots, s_{t-1}, s_t), \{o^\kappa\}_{\kappa = 1}^{\textbf{t}-1})$, where $\bar{t}$ denotes the time step at which the observation state $s$ was at the end of the last option $o^{\textbf{t}-1}$. 
Based on the generated option chain, the expert demonstration trajectory becomes option-expanded $\tau^E = \left( G, \{s_t, o_{\textbf{t}}, a_t\}|_{1 \leqslant t \leqslant T}^{\textbf{1} \leqslant \textbf{t} \leqslant \textbf{K}} \right)$, thus constituting an SMDP. Subsequently, the current appropriate skill\zj{s} are selected based on the task goal and observations of the robotic agent. We define the selected skill $o^{\textbf{t}}$ to last \zj{for} $H(\textbf{t})$ time steps, where $H(\textbf{t})$ is variable, \zj{calculated} based on the time difference between $o^{\textbf{t}}$ and $o^{\textbf{t-1}}$ in the generated option chain. It is worth noting that in the following equations, we all use $H(\textbf{t})$ to represent the duration of the selected cognitive skill $o^{\textbf{t}}$.
After $H(\textbf{t})$ time steps, the option chain is invoked again to select a new skill based on the observations of the robotic agent. The policy $\pi$ predicts the action of the robotic agent at each time step $t$, conditional on the state observation and activated current skill at that time step.
CasIL invokes and learns discrete and combinable options in the generated option chain, rather than continuous embedding, as this encourages reuse and combination of the skills potentially embedded in these options, thus transferring information from visual input to actual behaviors. 

\paragraph{Training CasIL}
In CasIL, an agent, supervised by human cognitive priors, first learns how to generate cognitive skills at each key decision-making step, and then adjusts its actions based on these generated skills. This dual learning process gives rise to a dual training architecture as shown in Fig.{~\ref{archi}}. The innovation of CasIL lies in the dual cognition-action imitation produced by the high-level component that generate cognitive skills (option chain) and the low-level components that perform operations according to the skills generated by the upper-layer.
The high-level cognition generator module of CasIL uses a memory-augmented architecture, LSTM \cite{graves2012long}, to embed cognition histories (i.e., the option chain), the Transformer encoder \cite{chen2021decision} to process task and cognition histories, and FiLM \cite{perez2018film} to fuse textual representations of task goals with visual observation inputs that have been aligned to human cognitive priors. The low-level policy learning component is identical to Behavior Cloning, except with the additional encoding of skills.
For a specific trajectory of length $T$ in the visual expert dataset, the training process minimizes the following objective function:
\begin{equation}
\begin{split}
        \mathcal{L}_{\text{CasIL}} = \text{min}_{\theta_g,\theta_p} \sum_{t=1}^T \big( -\varepsilon \underbrace{\text{log} \mathcal{F}_{\xi} (o^{\textbf{t}} | G, \{s_{\kappa}\}_{\kappa=1}^{\kappa=\sum H(\textbf{t})}, \{o^{\kappa}\}_{\kappa=\textbf{1}}^{\textbf{t}-1})}_{\textcolor{teal}{\textbf{Cogniton generator loss}}} \\
        -\underbrace{\text{log} \pi_{\theta} (\hat{a}_t|\hat{G},\{\hat{s}_{\kappa}\}_{\kappa=1}^{\kappa= \sum H(\textbf{t})}, o^{\textbf{t}})}_{\textcolor{olive}{\textbf{Policy learning loss}}}\big).
\end{split}
\label{eq1}
\end{equation}

Here, $\xi$ and $\theta$ represent the weights of the high-level cognition generator module $\mathcal{F}$ and the low-level policy learning module $\pi$, respectively. $\varepsilon$ denotes the coefficient of loss of cognitive generation. $o, s, a$ and $G$ denote options (skills), observations, actions, and task goals, respectively, as previously described. The pseudo-code for the training process of CasIL is shown in Algorithm{~\ref{alg:algorithm}}, and please refer to the appendix for more detailed training modules in CasIL.
\begin{algorithm}[tb]
\caption{Training CasIL}
\label{alg:algorithm}
\textbf{INPUT}: expert demonstration dataset $\mathbb{D} = \{\tau_i\}_{i=1}^N$, where $\tau_i = (G, \{s_t, a_t\}_{t=1}^T)$, textual descriptions of observations under critical decisions in the logic of human thinking $\{l_{\textbf{1}},\dots,l_{\textbf{K}}\}$\\
\textbf{INPUT}: Initialize high-level cognition generator $\mathcal{F}_{\xi}(o|s,G, \underset{\textbf{Option Chian}}{\underbrace{\{o^{\textbf{1}}, o^{\textbf{2}}, \dots\}}})$, 
low-level policy learning component $\pi_\theta(\hat{a}|\hat{s},\hat{G},o)$ 
\begin{algorithmic}[1] 
\While{\textit{not converged}}
\For {each $\tau_i = (G, \{s_t, a_t\}_{t=1}^T)$ in $\mathbb{D}$}
    \For {each $l_{\textbf{t}}$ in $\{l_{\textbf{1}},\dots,l_{\textbf{K}}\}$}
    \For {each $(G, \{s_t, a_t\})$ in $\tau_i$}
    \If {$s_t$ align with $l_{\textbf{t}}$}
    \State Generate high-level cognitive option chain $o^{\textbf{t}} = \mathcal{G}_\xi(\dot|G,\{s_{\kappa}\}_{\kappa=1}^{\kappa=t}, \{o^{\kappa}\}_{\kappa=1}^{\textbf{t}-1})$.
    \State Predict low-level action probability distribution $\hat{a_t} = \pi_{\theta}(\dot|\hat{G},\{\hat{s}_{\kappa}\}_{\kappa=1}^{\kappa=t},o^{\textbf{t}})$ 
    \EndIf
    \State Train \textcolor{teal}{cognition generator loss} $\mathcal{L}_{\mathcal{F}_\xi}$ and \textcolor{olive}{policy learning loss} $\mathcal{L}_{\pi_\theta}$ using objective $\mathcal{L}_{\text{CasIL}}$
    \State Break \textcolor{blue}{\Comment{Jump out of the current \textbf{for} loop}}
    \EndFor
    \EndFor 
\EndFor
\EndWhile
\end{algorithmic}
\end{algorithm}

\section{Experiments}
In this section we critically evaluate our architecture using widely recognized benchmarks for robot learning, demonstrating that CasIL enables robots to achieve highly competitive and robust skill imitation performance in a variety of long-horizon control tasks.
\subsection{Environment Setup}
\begin{figure}[htb!]
\centering
\includegraphics[width=0.75\columnwidth]{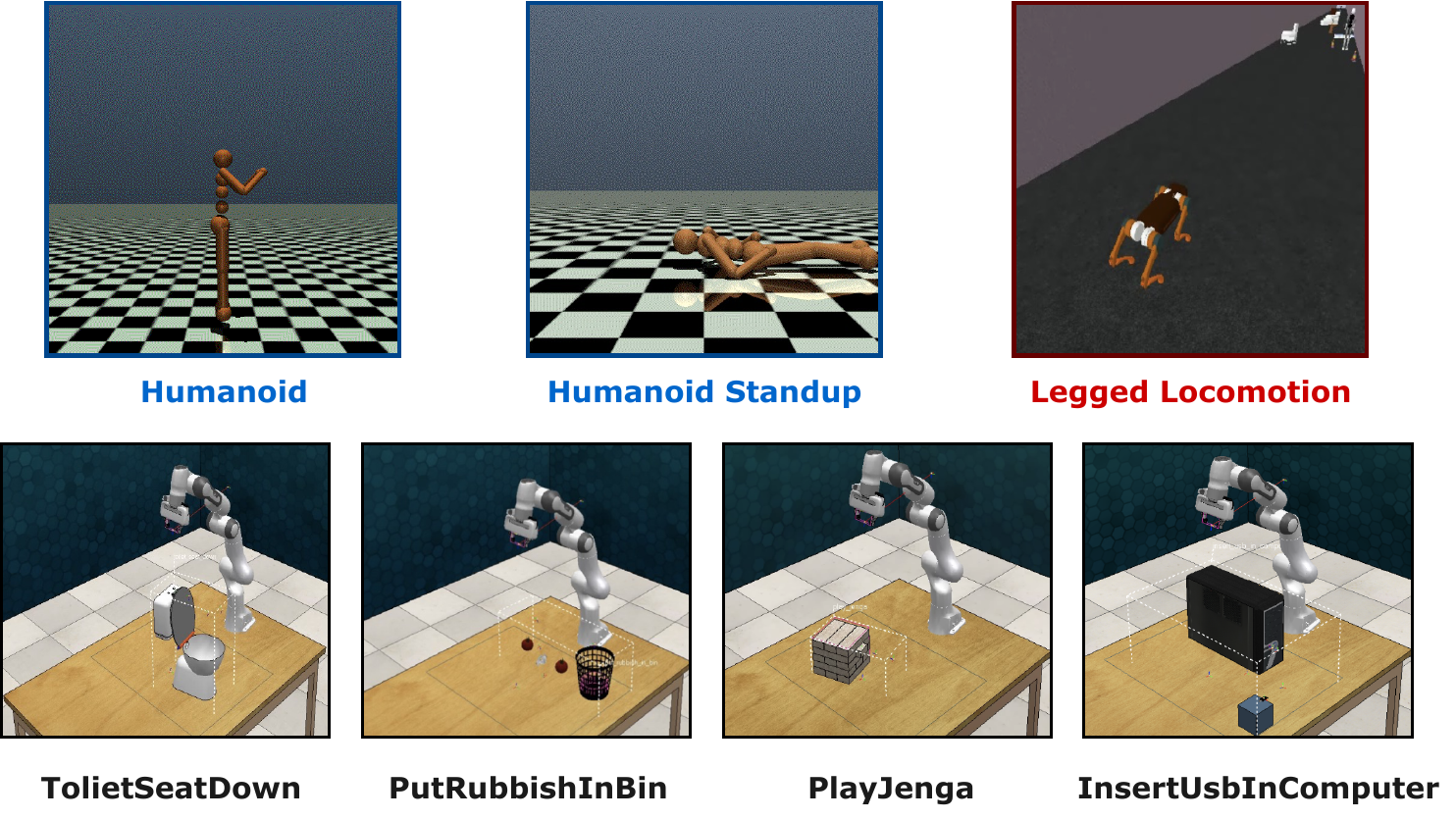} 
\caption{Robotic simulation environments.}
\label{simulation}
\end{figure}

Seven representative robotic simulation environments are used for evaluations as depicted in Fig.{~\ref{simulation}}.

\textbf{Humanoid Robot Locomotion.} The Humanoid-v2 and HumanoidStandup-v2 are two standardized continuous-time humanoid robot locomotion environments implemented in the OpenAI Gym \cite{brockman2016openai} based on the MuJoCo physics simulator \cite{todorov2012mujoco}. 
\zj{In} both environments, a 3D bipedal robot is designed to simulate a human. The goal of the Humanoid-v2 is to walk forward as fast as possible without falling over. 
\zj{The HumanoidStandup-v2 starts with the state of laying on the ground, and the goal is to enable the humanoid to learn to stand up and keep balanced by adaptively varying the torques on the joints.} 
We use expert demonstrations containing 2,000 time steps in the learning of both types of humanoid robot locomotion tasks. The expert demonstrations for imitation are generated from a well-trained PPO \cite{schulman2017proximal}. To quantitatively evaluate each model in both environments, we use the mean and standard deviation of the \textit {cumulative reward} tested in 5 new tasks with 5 random seeds per new task as the evaluation metrics.
\begin{table}[htb!]
\centering
\renewcommand\arraystretch{1.1}
\resizebox{\columnwidth}{!}{%
\begin{tabular}{cll}
\hline \hline
                                    & \multicolumn{1}{c}{Humanoid-v2} & \multicolumn{1}{c}{HumanoidStandup-v2} \\ \hline
Demos $(s,a) \times T$              & \multicolumn{1}{c}{($\mathbb{R}^{376},\mathbb{R}^{17}$) $\times$ 3k}            & \multicolumn{1}{c}{($\mathbb{R}^{376},\mathbb{R}^{17}$) $\times$ 2k}                   \\
Demo Cumulative Reward              &       130842.95 $\pm$ 28841.72                           &        137360.10 $\pm$ 30519.09                                \\ \hline
BC                                  & \multicolumn{1}{c}{1904.31 $\pm$ 584.29}           & \multicolumn{1}{c}{428.62 $\pm$ 39.10}                   \\
H-BC                                & \multicolumn{1}{c}{69661.03 $\pm$ 9929.44}                           &    \multicolumn{1}{c}{42360.16 $\pm$ 2008.32}                                     \\
Option-GAIL                         & \multicolumn{1}{c}{63517.91 $\pm$ 11824.95}                                &        \multicolumn{1}{c}{67617.31 $\pm$ 6356.68}                               \\
CasIL w/o Cognition                 & \multicolumn{1}{c}{72481.52 $\pm$ 21041.18}                                &      \multicolumn{1}{c}{74966.87 $\pm$ 6506.02}                                  \\ \hline
\rowcolor[HTML]{EFEFEF} 
{ \textbf{\texttt{CasIL} (ours)}} &    \multicolumn{1}{c}{ {\textbf{123814.10 $\pm$ 35981.58}}}                    &    \multicolumn{1}{c}{ {\textbf{114133.27 $\pm$ 42854.62}}}                                     \\ \hline \hline
\end{tabular}%
}
\caption{Comparison of test results for two humanoid robot control tasks on MuJoco benchmark. Better performance is indicated by being closer to the Demo Cumulative Reward.}
\label{tab:my-table1}
\end{table}

\textbf{Robotic Arm Manipulation.} We conducted experiments on RLbench \cite{james2020rlbench} for the evaluation of the architecture in robotic arm manipulation tasks using the four settings of increasing manipulation difficulty (increasing task horizon and uncertainty in the environment), with a total of 100 demonstration trajectories for IL training in each setting. The robotic agent in each of the following setups is a 6-DOF robotic arm with a gripper, which performs the object manipulation task by moving its joints.
\textbf{ToiletSeatDown:} In this setting, the robotic arm is required to grip the edge of the toilet lid and lower it flat on the toilet seat within a defined maximum of 200 time steps. 
\textbf{PutRubbishInBin:} In this setting, the robotic arm is required to pick up the rubbish and drop all the rubbish into the bin within a defined maximum 250 time-steps.
\textbf{PlayJenga:} In this setting, the robotic arm with is required to take the protruding blocking out of the jenga tower without the tower topping within a defined maximum 300 time-steps.
\textbf{InsertUsbInComputer:} In this setting, the robotic arm is required to pick up the Usb stick and insert it into the Usb port, within a defined maximum of 400 time steps.  Across all tasks, we quantitatively evaluate each model using the mean and standard deviation of the \textit{success rates} tested on 80 randomly initialized new scenarios as the primary metric.

\textbf{Quadrupedal Robot Locomotion.} The point-goal navigated quadrupedal robot locomotion obstacle avoidance task employs a corridor simulation environment as proposed by \cite{seo2023learning} with a width of 3 meters width and 3 meters length, 
\zj{based on the} Bullet Physics \zj{engine}~\cite{coumans2015bullet}. The environment 
\zj{i}s initialized with a scenario containing varying numbers of static obstacles and walking humans, and the task 
\zj{i}s categorized into three difficulty levels with a total of 80 demonstration trajectories 
each level: 1) Easy level: Corridor environment with sparse obstacles (average of 5) and only 0 or 1 moving human; 2) Medium level: Cluttered environment with an average of 15 static obstacles and 0 or 1 moving human; \zj{and} 3) Hard level: Crowded environment with an average of 15 static obstacles and 0 or 1 moving human. The more objects in the corridor, the greater the likelihood of robot collisions in each difficulty level. In all tasks, the footed robot succeeds in the task when it reached the goal point from the starting point without collision.  To quantitatively evaluate each model, we use the \textit{average traverse length in meters} (total length of the corridor: 50 meters) and the \textit{average success rate} tested in 80 randomly initialized new scenes as the basic metric for imitation performance evaluation. 

\begin{table*}[]
\centering
\renewcommand{\arraystretch}{1.05}
\resizebox{\textwidth}{!}{%
\begin{tabular}{ccccc|ccc}
\hline \hline
\multicolumn{1}{l}{} &
  \multicolumn{4}{c|}{\textbf{Robotic Arm Manipulation}} &
  \multicolumn{3}{c}{\textbf{Quadrupedal Robot Locomotion}} \\ \hline
\multicolumn{1}{l}{} &
  \multicolumn{1}{c|}{\textit{ToiletSeatDown}} &
  \multicolumn{1}{c|}{\textit{PutRubbishInBin}} &
  \multicolumn{1}{c|}{\textit{PlayJenga}} &
  \textit{InsertUsbInComputer} &
  \multicolumn{1}{c|}{Easy Level} &
  \multicolumn{1}{c|}{Medium Level} &
  Hard Level \\ \hline
\multicolumn{1}{l}{} &
  \multicolumn{4}{c|}{\textit{\textbf{Success Rate (in \%)}}} &
  \multicolumn{3}{c}{\textit{\textbf{Average Length (Success Rate (in \%))}}} \\ \hline
BC &
  93.7 $\pm$ 4.3 &
  74.4 $\pm$ 3.7 &
  21.5 $\pm$ 8.8 &
  00.0 $\pm$ 0.0 &
  12.3 $\pm$ 8.3 (3\%) &
  10.1 $\pm$ 3.4 (1\%) &
  3.3 $\pm$ 1.5 (0\%) \\
H-BC &
  98.5 $\pm$ 1.5 &
  85.2 $\pm$ 5.9 &
  33.6 $\pm$ 7.9 &
  10.6 $\pm$ 1.8 &
  26.8 $\pm$ 11.2 (34\%) &
  22.1 $\pm$ 14.7 (29\%) &
  16.0 $\pm$ 6.2 (21\%) \\
Option-GAIL &
  99.0 $\pm$ 1.0 &
  81.4 $\pm$ 9.6 &
  48.2 $\pm$ 9.2 &
  23.3 $\pm$ 5.5 &
  30.1 $\pm$ 17.7 (47\%) &
  26.2 $\pm$ 16.6 (33\%) &
  21.4 $\pm$ 8.7 (28\%) \\
CasIL w/o Cognition &
  99.4 $\pm$ 0.6 &
  89.6 $\pm$ 9.4 &
  53.1 $\pm$ 8.3 &
  26.4 $\pm$ 4.1 &
  34.4 $\pm$ 16.3 (64\%) &
  32.0 $\pm$ 13.2 (48\%) &
  25.6 $\pm$ 5.3 (32\%) \\ \hline
\rowcolor[HTML]{ECF4FF} 
\cellcolor[HTML]{EFEFEF}{ \textbf{\texttt{CasIL} (ours)}} &
  { \textbf{100.0 $\pm$ 0.0}} &
  { \textbf{98.4 $\pm$ 1.6}} &
  { \textbf{82.4 $\pm$ 3.5}} &
  { \textbf{57.6 $\pm$ 2.4}} &
  \cellcolor[HTML]{FEECEB}{ \textbf{46.2 $\pm$ 12.8 (87\%)}} &
  \cellcolor[HTML]{FEECEB}{ \textbf{44.2 $\pm$ 14.4 (81\%)}} &
  \cellcolor[HTML]{FEECEB}{ \textbf{36.3 $\pm$ 18.4 (65\%)}} \\ \hline \hline
\end{tabular}%
}
\caption{Comparison of test results under four robotic arm manipulation tasks (left side) and three quadruped robot obstacle avoidance difficulty tasks (right side). Larger values indicate better performance.}
\label{tab:my-table2}
\end{table*}

\begin{figure}[htp!]
\centering
\subfigure[]{
    \includegraphics[width=3.5cm]{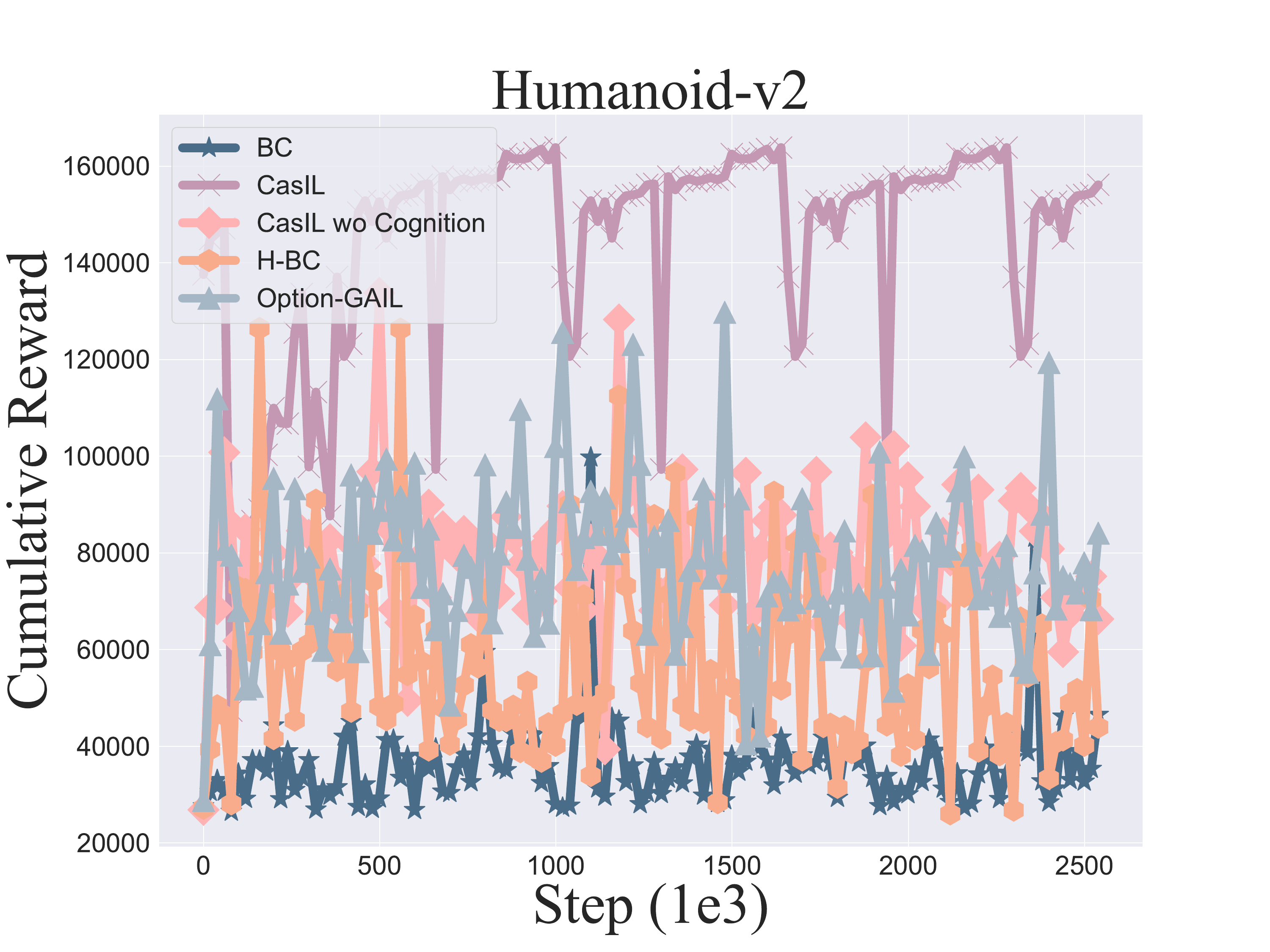}
           \label{humanoid:1}
           }
\subfigure[]{
    \includegraphics[width=3.5cm]{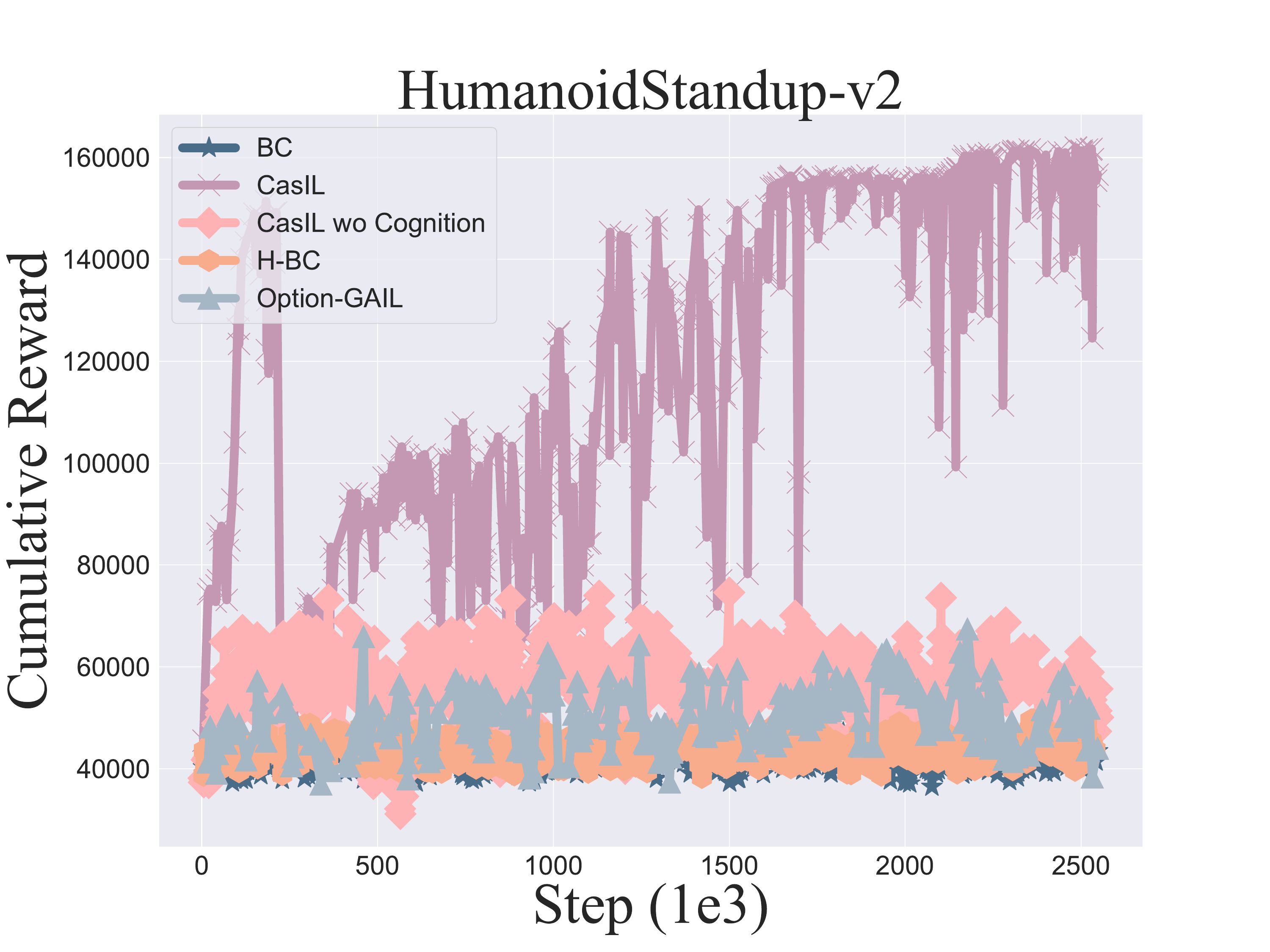}
           \label{humanoid:2}
           }
\caption{Comparison of training performance for two humanoid robot control tasks on MuJoco benchmark.}
\label{humanoid}
\end{figure}

\begin{figure}[htb!]
\centering
\includegraphics[width=0.7\columnwidth]{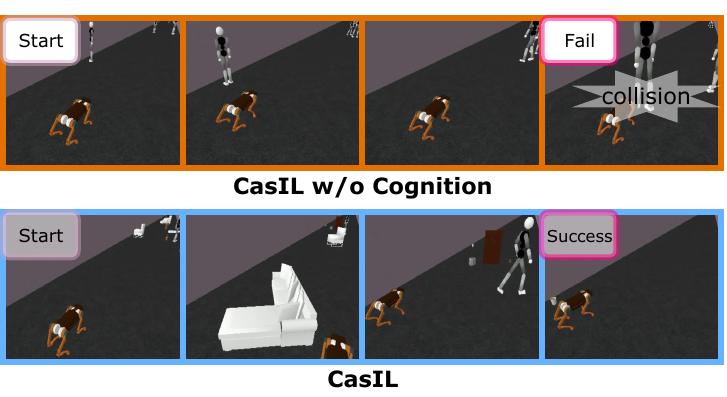} 
\caption{Visualized test instances of ablation experiments for ``hard" level obstacle avoidance tasks.}
\label{locomotion}
\end{figure}

\begin{table*}[]\small
\centering
\renewcommand{\arraystretch}{1.3}
\Huge
\resizebox{\textwidth}{!}{%
\begin{tabular}{c|ccc|ccc|ccc|ccc}
\hline \hline
 &
  \multicolumn{3}{c|}{\textit{ToiletSeatDown}} &
  \multicolumn{3}{c|}{\textit{PutRubbishInBin}} &
  \multicolumn{3}{c|}{\textit{PlayJenga}} &
  \multicolumn{3}{c}{\textit{InsertUsbInComputer}} \\ \cline{2-13}
\multirow{-2}{*}{\textbf{Expert Demo Trajs Num}} &
  \multicolumn{1}{c|}{H-BC} &
  \multicolumn{1}{c|}{CasIL w/o Cognition} &
  \cellcolor[HTML]{E9EAF5}\textbf{\texttt{CasIL} (ours)} &
  \multicolumn{1}{c|}{H-BC} &
  \multicolumn{1}{c|}{CasIL w/o Cognition} &
  \cellcolor[HTML]{E9EAF5}\textbf{\texttt{CasIL} (ours)} &
  \multicolumn{1}{c|}{H-BC} &
  \multicolumn{1}{c|}{CasIL w/o Cognition} &
  \cellcolor[HTML]{E9EAF5}\textbf{\texttt{CasIL} (ours)} &
  \multicolumn{1}{c|}{H-BC} &
  \multicolumn{1}{c|}{CasIL w/o Cognition} &
  \cellcolor[HTML]{E9EAF5}\textbf{\texttt{CasIL} (ours)} \\ \hline
 &
  \multicolumn{3}{c|}{\textit{\textbf{Success Rate (in \%)}}} &
  \multicolumn{3}{c|}{\textit{\textbf{Success Rate (in \%)}}} &
  \multicolumn{3}{c|}{\textit{\textbf{Success Rate (in \%)}}} &
  \multicolumn{3}{c}{\textit{\textbf{Success Rate (in \%)}}} \\ \hline
20 trajs (80\% data drop $\downarrow$) &
  70.8 $\pm$ 6.7 &
  81.3 $\pm$ 5.2 &
  \cellcolor[HTML]{E9EAF5}\textbf{99.4 $\pm$ 0.6} &
  52.5 $\pm$ 6.6 &
  60.4 $\pm$ 6.7 &
  \cellcolor[HTML]{E9EAF5}\textbf{89.4 $\pm$ 1.7} &
  8.7 $\pm$ 0.8 &
  19.8 $\pm$ 0.9 &
  \cellcolor[HTML]{E9EAF5}\textbf{73.8 $\pm$ 2.3} &
  0.0 $\pm$ 0.0 &
  0.0 $\pm$ 0.0 &
  \cellcolor[HTML]{E9EAF5}\textbf{41.9 $\pm$ 1.6} \\ \cline{1-1}
50 trajs (50\% data drop $\downarrow$)&
  77.8 $\pm$ 8.5 &
  89.7 $\pm$ 3.5 &
  \cellcolor[HTML]{E9EAF5}\textbf{100.0 $\pm$ 0.0} &
  61.8 $\pm$ 3.1 &
  73.3 $\pm$ 5.3 &
  \cellcolor[HTML]{E9EAF5}\textbf{92.7 $\pm$ 2.6} &
  17.4 $\pm$ 1.7 &
  32.1 $\pm$ 2.6 &
  \cellcolor[HTML]{E9EAF5}\textbf{77.4 $\pm$ 3.1} &
  0.0 $\pm$ 0.0 &
  5.3 $\pm$ 1.5 &
  \cellcolor[HTML]{E9EAF5}\textbf{48.3 $\pm$ 2.7} \\ \cline{1-1}
80 trajs (20\% data drop $\downarrow$)&
  95.8 $\pm$ 1.1 &
  97.2 $\pm$ 0.8 &
  \cellcolor[HTML]{E9EAF5}\textbf{100.0 $\pm$ 0.0} &
  79.0 $\pm$ 8.3 &
  82.4 $\pm$ 2.1 &
  \cellcolor[HTML]{E9EAF5}\textbf{96.3 $\pm$ 1.2} &
  25.4 $\pm$ 1.1 &
  44.2 $\pm$ 4.6 &
  \cellcolor[HTML]{E9EAF5}\textbf{80.2 $\pm$ 2.2} &
  3.1 $\pm$ 0.4 &
  19.4 $\pm$ 2.4 &
  \cellcolor[HTML]{E9EAF5}\textbf{51.4 $\pm$ 3.1} \\ \hline \hline
\end{tabular}%
}
\caption{Test performance of models trained using different numbers of demo trajectories in robotic arm manipulation tasks.}
\label{tab:my-table3}
\end{table*}

\subsection{Comparative Evaluations}
To give a fairer illustration of the effectiveness of the proposed architecture, we compare several typical IL baselines, including 1) \textbf{supervised behavioural cloning (BC)} \cite{pomerleau1988alvinn}, which does not contain any hierarchical structure or decision-cognitive modules; 2) \textbf{hierarchical behavioural cloning (H-BC)} \cite{zhang2021provable}, a HBC-type baseline built on an option architecture but optimizes high- and low-level polices by directly maximizing the probability of observed expert trajectories, and does not incorporate the agent's self-exploration of the hierarchy; 3) \textbf{Option-GAIL} \cite{jing2021adversarial}, a HIRL-type baseline that builds on a one-step option architecture with self-exploration by agent to generate hierarchical task decisions, but without the guidance of human cognitive prioris. 4) The ablation variant \textbf{CasIL w/o Cognition}, which has no cognition generator loss during training, thus demonstrating that the superiority of CasIL does not lie solely in its higher number of parameters. Latent embeddings in the high-level of this variant are directly fed into the low-level as skills through only a one-time alignment of human cognitive text and expert demonstration images.
For a fair comparison, we set the number of options in H-BC and Option-GAIL in each task to match the optimal number of skills in CasIL.  
All of the primary results in this study were achieved in a month on four V100 GPUs, including the collection of expert demonstrations. Please refer to the Appendix for more training details. 



As shown in Fig.{~\ref{humanoid}}, in the two humanoid robot control experiments using the MuJoCo benchmark, Humanoid-v2 and HumanoidStandUp-v2 (both of which have significantly long horizons), CasIL outperforms all baseline methods in terms of the final cumulative reward during training. Additionally, Option-GAIL, H-BC, and CasIL w/o Cognition outperform comparable schemes utilizing a single-layer policy, i.e., BC, by introducing an option-based hierarchical structure. Furthermore, the results in Table{~\ref{tab:my-table1}} intuitively show that in the test scenarios of the two 
humanoid robot control tasks, the robot trained with CasIL achieves the best imitation performance, i.e., it obtains a cumulative reward closest to that of the expert demonstration. 

Comparison results in the robotic arm manipulation task (left side of Table{~\ref{tab:my-table2}}) show that all methods achieve good skill imitation performance on the simplest \textit{ToiletSeatDown} task, where our CasIL achieves a 100\% success rate on all test tasks. However, as the difficulty of the task increases (the more objects, the worse the stability), the success rate of BC's skill imitation drops off a cliff, even 0\% in all test tasks of \textit{InsertUsbInComputer}. 
H-BC and Option-GAIL, baselines without a supervision of human cognitive priors, fall far short of the skill imitation performance of our CasIL, while CasIL w/o Cognition also fails to maintain stable manipulation imitation due to the absence of continuous training on text-image alignment.
Notably, the performance of Option-GAIL suggests that one-step option architecture based on agent self-exploration is unable to achieve stable skill imitation in complex long-horizon tasks. 

In addition, to validate the robustness of CasIL in low-expert-data regime, we train with varying numbers of expert demo trajectories and test the trained model on 80 new scenarios. Comparison of the test results is shown in Table{~\ref{tab:my-table3}}. It can be clearly seen that with the reduction of expert data, our CasIL does not suffer from the dramatic drop in imitation performance 
\zj{as with the} H-BC and CasIL w/o Cognition baselines, 
and can still perform relatively stable manipulation.
\zj{This} is attributed to the cognition generator in CasIL that interacts with human cognition, providing guidance that is \zj{considerably stable} 
even in the low-data regime. 
In the obstacle avoidance tasks with the quadrupedal robot, the comparative results on the right side of Table{~\ref{tab:my-table2}} show that CasIL achieves the best performance in all three difficulty levels. 
The failure of the BC baseline shows that all three levels of environments require the robot to hierarchically learn flexible collision avoidance behaviors. 
While the results of the hierarchical baseline show that robots can, to some extent, walk in simple static environments by decomposing tasks based on their exploration experience. However, it is difficult to imitate complex obstacle avoidance navigation skills based on the robot's own experience alone, especially when dealing with dynamic obstacles. 
The results of H-BC and Option-GAIL show that the lack of human cognitive priors leads to a decline in the imitation performance of obstacle avoidance skills, while the results of CasIL w/o Cognition show that, even when the human priors are utilized, the lack of a cognition generator for the text-image-aligned training renders the generated skills unreliable, especially at the ``hard" level, and our final CasIL model walks an additional 33\% of the average distance, demonstrates the robustness of CasIL in imitating complex obstacle-avoidance skills in dynamic environments.
Fig.{~\ref{locomotion}} visualizes a comparison of test instances of CasIL w/o Cognition and CasIL in the ``Hard" level of obstacle avoidance. We observed that CasIL w/o Cognition trained the robot to drift violently after a high-speed turn and hit a moving person, whereas, CasIL trained the robot to make a quick turn around a walking crowd and an obstacle and successfully completed the trial. This means that by training the cognition generator's text-image-aligned process, CasIL continuously refines the generation of the robot's cognition through complete human-robot interaction, leading to a more stable imitation of obstacle avoidance skills in dynamic environments. We have provided video demos in the Multimedia Material.

\section{Conclusion and Discussion}
We present CasIL, a dual cognition-action imitation architecture for robotic skill learning, which learns text-image-alignment cognition from visual demonstrations inspired by human cognitive priors, thus enabling robots to cognize and imitate skill abstractions for critical decisions.
Experimental results demonstrate that CasIL enables robotic agents to perform superior and robust skill imitation in diverse control tasks such as humanoid locomotion, manipulation, and obstacle avoidance, even in tasks with low-expert-data.

Nevertheless, there remains intriguing challenges that deserve further research. For example, human cognitive priors provide only a limited set of skills and a limited time span for each skill, which may impair agent's ability to consistently imitate and generalize skills across multi-task outside the distribution of training demonstrations. Addressing these issues is an important goal for our ongoing and future work. 

\renewcommand*{\thesection}{\Alph{section}}
\renewcommand*{\thesubsection}{\thesection.\arabic{subsection}.}

\bibliography{aaai24}

\clearpage
\appendix
\section*{Technical Appendix}
In this technical appendix, we first summarize the detailed notations used in the ``Method" section, and present details about the CasIL architecture.
Then we describe our experimental details including the  information on the dataset, open source libraries used in our architecture, and the hyperparameters. 
We further provide additional ablations and more experimental results. Finally, we provide more details on the preliminaries.
\begin{figure*}[t]
\centering
\includegraphics[width=\textwidth]{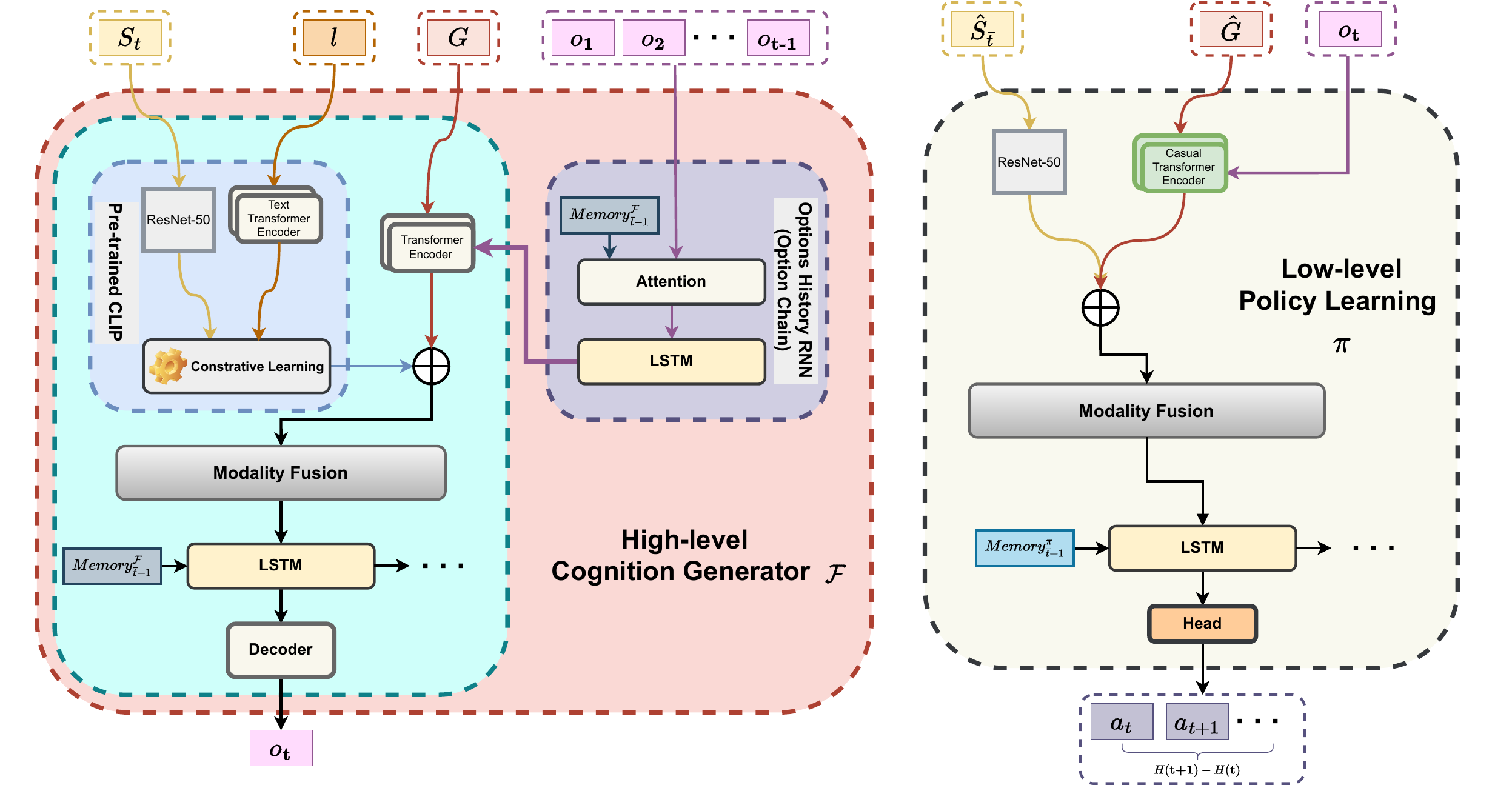} 
\caption{\textbf{Detailed CasIL Architecture.} The preceding option chain $\{o_{\textbf{1}},o_{\textbf{2}},\cdots,o_{\textbf{t-1}} \}$ is embedded with an LSTM in the High-level Component. The generated skill (option) $o_{\textbf{t}}$ from the High-level Component will be the input to the Low-level Component and a sequence of actions are predicted by the Low-level Component. (\textbf{Left: High-level Component.)} We employ an LSTM \cite{graves2012long} to embed the option chain, a pretrained CLIP \cite{radford2021learning} to align the human cognitive priors with visual observation, and a transformer encoder \cite{vaswani2017attention} to process both the expert task goal and option chain. The text input of task goal is then fused with the aligned visual observation using FiLM \cite{perez2018film}. (\textbf{Right: Low-level Component.)} The Low-level Component is identical to the Behavior Cloning, except with the additional encoding of skills (options).}
\label{details}
\end{figure*}

\subsection{Notations}  Table{~\ref{tab:notation}} summarizes the detailed notations used in the ``Method'' section.

\begin{table}[h!]
\centering
\begin{tabular}{c|c}
\toprule
    Symbol & Meaning  \\
    \midrule
    $s$ & The observation (state) of agent  \\
    $a$ & The action of agent \\
    $\mathcal{S}$ & The observation (state) set of agent \\
    $\mathcal{A}$ & The action set of agent \\
    $o$ & An option of agent\\
    $\mathcal{O}$ & The option chain of agent \\
    $\pi$ & The policy of agent \\
    $\mathcal{P}$ & The transition function of agent\\
    $\mathcal{R}$ & The reward function of agent\\
    $\mathcal{I}$ & The initiation state set of an option \\
    $\beta$ & The termination condition of an option \\
    $\mathcal{T}$ & A long-horizon task \\
    $\mathbb{D}$ & An offline dataset of task $\mathcal{T}$   \\
    $\tau$ & A trajectory in $\mathbb{D}$ \\
    $G$ & The task goal of a long-horizon task \\
    $\{\textbf{1},\cdots,\textbf{K}\}$ & The number of options in an option chain $S$ \\
    $\bar{t}$ & The critical time steps to start an option  \\
    $H({\textbf{t}})$ & The duration of option $o^{\textbf{t}}$\\
\bottomrule
\end{tabular}
\caption{Notation for Method Section}\label{tab:notation}
\end{table}

\subsection{CasIL Implementation Details}
\paragraph{High-level Components} The architectural details of CasIL architecture are shown in Fig.{~\ref{details}}. All task information and options are encoded with Gated Linear Units (GLUs). In the high-level component, we apply an attention mechanism \cite{vaswani2017attention} to dynamically weight the importance of different parts of the option chain, based on the expert observation history. To achieve the alignment of textual forms of human cognitive priors with expert visual observations in Cognition Generator, we use a pre-trained CLIP \cite{radford2021learning} for contrastive learning, according to the cosine similarity of text and image features, the visual embeddings with the highest similarity to text features are selected as the critical decision states for the cognitive generation. Specifically, we use a pre-trained text transformer encoder \cite{vaswani2017attention,radford2019language} to generate language embeddings from human cognitive textual description inputs. For visual observations, we use ResNet-50 \cite{he2016deep} to generate visual embeddings.

After the alignment process, in the high-level cognition generator component, we use a Transformer encoder \cite{vaswani2017attention} to embed cognitive skills and sub-task goals, with cognitive skills as queries and sub-task goals as keys and values. 

\paragraph{Low-level Components}The low-level policy network component inspired by the Decision Transformer (DT) \cite{chen2021decision}, implemented as a small causal Transformer with additional cognitive skill encoding, and the sequence length of generated policy is the horizon $H(\textbf{t})$ of the skill $o^{\textbf{t}}$ which is much smaller compared to the length of the full trajectory.

\subsection{Training Details}
\paragraph{Dataset}
We conducted experiments on three types of robot control tasks, and the following details the datasets we used in our experiments, including how the expert demonstrations for each type of task were collected and the associated codebase.
\begin{itemize}
    \item \textbf{Humanoid Robot Locomotion.} In this experiment, we used the Humanoid-v2 and HumanoidStandup-v2 , which are two standardized continuous-time humanoid robot locomotion environments implemented in the OpenAI Gym\footnote{https://github.com/openai/gym} \cite{brockman2016openai} based on the MuJoCo physics simulator\footnote{https://github.com/deepmind/mujoco}\cite{todorov2012mujoco}. We use expert demonstrations containing 2,000 time steps in the learning of both types of humanoid robot locomotion tasks. The expert demonstrations for imitation are generated from a well-trained PPO\footnote{https://github.com/openai/baselines} \cite{schulman2017proximal}. In both Humanoid-v2 and HumanoidStandup-v2, the robot's action space is 17 dimensions and the observation space is 376 dimensions, an action represents the torques applied at the hinge joints. 
    \item \textbf{Robotic Arm Manipulation.} In this experiment, we evaluate on 4 available RLBench\footnote{https://github.com/stepjam/RLBench/tree/master/rlbench/tasks} \cite{james2020rlbench} tasks: \textit{ToiletSeatDown}, \textit{PutRubbishInBin}, \textit{PlayJenga}, and \textit{InsertUsbInComputer}. Each task is accompanied by a series of textual descriptions that verbally summarize the task and are therefore well suited to the human-robot interaction portion of our CasIL. Expert demonstrations in the four tasks were collected through the Open Exercise Planning Library \cite{sucan2012open}.
    \item \textbf{Quadrupedal Robot Locomotion.} In this experiment, we used a simulated corridor environment\footnote{https://github.com/UT-Austin-RPL/PRELUDE} in Bullet Physics~\cite{coumans2015bullet} with 3 meters width and 50 meters length that proposed by~\citet{seo2023learning}. We utilized this environment to systematically compare the ability of our approach and a comparison baseline to imitate obstacle avoidance skills in a dynamic environment under three difficulty conditions. As with \cite{seo2023learning}, all static obstacles in this experimental simulation environment were initialized with 3D assets from ShapeNet \cite{chang2015shapenet} and Google Scanned Objects \cite{downs2022google}, which cover everyday objects. Human motion is controlled by a motion generation method based on Gaussian processes. In addition, random external perturbations were added to the robot to simulate the environmental uncertainty that often exists in the real world. In addition, the expert demonstration trajectories used in the experiments were obtained from the \cite{seo2023learning} open source data\footnote{https://utexas.app.box.com/s/vuneto210i5o5c8vi09cxt49dta2may3}, with 80 trajectories in each difficulty level were collected by a human pushing a cart equipped with an Intel RealSense D435 camera, a RealSense T265 stereo camera, and the VectorNav VN100 IMU sensor. Each trajectory containing information: 1) observation data, including RGB-D images and the heading orientationl; 2) the forward linear and angular velocity of the cart. It is worth noting that in this experiment we only imitate the obstacle avoidance skills of robot navigation, i.e., the forward and angular velocities of the robot, while the gait controllers are always utilized with the pre-trained ppo policy in \cite{seo2023learning}.
\end{itemize}

\paragraph{Hyperparameters}
Detailed hyperparameter settings are shown in Table{~\ref{tab:hyper}}. The Adam optimizer is apdoted in the training process, with a batch size of 180 and a learning rate of $5e^{-4}$. The learning rate schedule begins with a warm-up phase of $10$ training steps, linearly increasing from $1e^{-4}$ to $5e^{-4}$, and then decaying by 50\% at 150th training steps. For training efficiency, Backpropagation Through Time was truncated at 30th training steps in CasIL. The mix precision in PyTorch is also adopted during training, which speeds up training without sacrificing much performance. 

\begin{table*}[h!]
\centering
\begin{tabular}{c|c}
\toprule
    \textbf{Hyperparameter} & \textbf{Value}  \\
    \midrule
    Image Embedding Dimension & 128 \\
    Text Embedding Dimension & 256 \\
    Memory Dimension & 2048 \\
    Transformer Layers & 2 \\
    Transformer Embedding Dim & 128\\
    Transformer Heads & 4 \\
    Policy Learning Rate & $5e^{-4}$ \\
    Dropout & 0.1 \\
    Batch Size & 180 \\
    Number of Human Cognitive Skills (Humanoid-v2) & 6 \\
    Number of Human Cognitive Skills (HumanoidStandip-v2) & 8 \\
    Number of Human Cognitive Skills (ToiletSeatDown) & 3 \\
    Number of Human Cognitive Skills (PutRubbishInBin) & 4 \\
    Number of Human Cognitive Skills (PlayJenga) & 7 \\
    Number of Human Cognitive Skills (InsertUsbInComputer) & 8 \\
    Number of Human Cognitive Skills (Obstacle Avoidance) & 12 \\

\bottomrule
\end{tabular}
\caption{Hyperparemeters for Experiments Section}\label{tab:hyper}
\end{table*}

\begin{table*}[t!]
\centering
\renewcommand{\arraystretch}{1.8}
\Huge
\resizebox{\textwidth}{!}{%
\begin{tabular}{l|ccccccccc}
\hline \hline
 &
  \multicolumn{9}{c}{\textbf{Legged Locomotion}} \\ \hline
 &
  \multicolumn{3}{c|}{Easy Level} &
  \multicolumn{3}{c|}{Medium Level} &
  \multicolumn{3}{c}{Hard Level} \\ \hline
 &
  \multicolumn{1}{c|}{20 expert trajs (75\% $\downarrow$)} &
  \multicolumn{1}{c|}{40 expert trajs (50\% $\downarrow$)} &
  \multicolumn{1}{c|}{60 expert trajs (25\% $\downarrow$)} &
  \multicolumn{1}{c|}{20 expert trajs (75\% $\downarrow$)} &
  \multicolumn{1}{c|}{40 expert trajs (50\% $\downarrow$)} &
  \multicolumn{1}{c|}{60 expert trajs (25\% $\downarrow$)} &
  \multicolumn{1}{c|}{20 expert trajs (75\% $\downarrow$)} &
  \multicolumn{1}{c|}{40 expert trajs (50\% $\downarrow$)} &
  60 expert trajs(25\% $\downarrow$)\\ \cline{2-10} 
 &
  \multicolumn{9}{c}{\textit{\textbf{Average Length (Success Rate (\%))}}} \\ \hline
\multicolumn{1}{c|}{H-BC} &
  8.2 $\pm$ 0.2 (4\%) &
  18.1 $\pm$ 2.3 (17\%) &
  \multicolumn{1}{c|}{21.6 $\pm$ 4.1 (25\%)} &
  9.7 $\pm$ 0.3 (2\%) &
  13.2 $\pm$ 1.8 (11\%) &
  \multicolumn{1}{c|}{9.7 $\pm$ 0.3 (2\%)} &
  2.0 $\pm$ 0.1 (0\%) &
  10.2 $\pm$ 1.1 (2\%) &
  13.0 $\pm$ 3.2 (11\%) \\
\multicolumn{1}{c|}{CasIL w/o Cognition} &
  16.4 $\pm$ 2.3 (13\%) &
  22.1 $\pm$ 4.2 (33\%) &
  \multicolumn{1}{c|}{32.9 $\pm$ 6.3 (48\%)} &
  15.7 $\pm$ 2.2 (9\%) &
  22.1 $\pm$ 3.4 (18\%) &
  \multicolumn{1}{c|}{15.7 $\pm$ 2.2 (9\%)} &
  8.2 $\pm$ 1.0 (5\%) &
  17.6 $\pm$ 1.9 (15\%) &
  20.9 $\pm$ 3.1 (29\%) \\ \hline
\rowcolor[HTML]{DAE8FC} 
\multicolumn{1}{c|}{\cellcolor[HTML]{DAE8FC}{\color[HTML]{9A0000} \textbf{CasIL (ours)}}} &
  {\color[HTML]{333333} \textbf{32.3 $\pm$ 7.7 (52\%)}} &
  \multicolumn{1}{l}{\cellcolor[HTML]{DAE8FC}{\color[HTML]{333333} \textbf{38.1 $\pm$ 9.8 (66\%)}}} &
  \multicolumn{1}{l|}{\cellcolor[HTML]{DAE8FC}{\color[HTML]{333333} \textbf{43.2 $\pm$ 10.3 (78\%)}}} &
  {\color[HTML]{333333} \textbf{30.7 $\pm$ 5.3 (48\%)}} &
  \cellcolor[HTML]{DAE8FC}{\color[HTML]{333333} \textbf{35.8 $\pm$ 6.1 (64\%)}} &
  \multicolumn{1}{c|}{\cellcolor[HTML]{DAE8FC}{\color[HTML]{333333} \textbf{40.1 $\pm$ 8.9 (75\%)}}} &
  {\color[HTML]{333333} \textbf{23.9 $\pm$ 3.7 (29\%)}} &
  \cellcolor[HTML]{DAE8FC}{\color[HTML]{333333} \textbf{27.3 $\pm$ 6.4 (41\%)}} &
  \cellcolor[HTML]{DAE8FC}{\color[HTML]{333333} \textbf{30.8 $\pm$ 9.7 (55\%)}} \\ \hline \hline
\end{tabular}%
}
\caption{Comparison of test performance in the quadrupedal robot obstacle avoidance task with varying amounts of expert data.}
\label{tab:appendix}
\end{table*}

\paragraph{Human Cognitive Priors for Each Task}
\begin{itemize}
    \item \textbf{Humanoid-v2}:     \textit{``Adjust gait for maximum speed."
    ``Adapt foot placement based on terrain feedback."
    ``Correct for imbalances in real-time."
    ``Prioritize safety and stability over speed."
    ``Monitor power reserves continuously."
    ``Stop if obstacle detection suggests high collision risk."}
    \item  \textbf{HumanoidStandup-v2}:  \textit{``Use limbs and actuators for leverage."
    ``Transition to a crouched stance."
    ``Rise to an upright position, adjusting center of gravity."
    ``Distribute weight evenly across feet."
    ``Monitor feedback from foot sensors for weight distribution."
    ``Make real-time micro-adjustments for upright stability."
    ``Ensure adequate power levels during this operation."
    ``Prioritize safety to prevent falls."}
    \item  \textbf{ToiletSeatDown}: \textit{``Grasp the toilet lid.''}, \textit{``Move the toilet lid downward.''}, \textit{``Release the toilet lid.''}
    \item  \textbf{PutRubbishInBin}: \textit{``Find rubbish."}, \textit{``Pick up rubbish."},\textit{``Move to the bin."}, \textit{``Throw rubbish in bin."}
    \item \textbf{PlayJenga}: \textit{``Identify loose blocks using light pressure tests."
``Opt for center blocks when available."
``Apply steady, slow force when extracting a block."
``Retract arm smoothly post-extraction."
``Align block to maintain tower balance."
``Continuously analyze tower stability."}
``Pause operation if risk of collapse is detected."
    \item \textbf{InsertUsbInComputer}: \textit{``Pick up the usb with gripper"
``Align arm orientation to the computer's plane."
``Approach socket slowly."
``Confirm USB orientation (typically, the USB logo facing up)."
``Adjust USB position for socket alignment."
``Insert gently until fully seated."
``Verify successful connection via force feedback."
``Maintain stable arm position post-insertion."}
    \item \textbf{Quadrupedal Robot Locomotion}:     \textit{``Set the corridor's end as your target."
    ``Scan surroundings using all sensors."
    ``Distinguish between static and dynamic obstacles."
    ``Prioritize walking humans and predict their paths."
    ``Engage stability algorithms; adjust using leg actuators."
    ``Navigate around static obstacles efficiently."
    ``Slow down near humans; maintain a safe distance."
    ``Monitor floor conditions for balance."
    ``Update your internal map continuously."
    ``Check power reserves regularly."
    ``Prioritize safety; re-evaluate if faced with high-risk obstacles."
    ``Proceed until you reach the goal point."}
\end{itemize}

\subsection{Supplementary Results}
\subsubsection{Robustness Validation}
We also conducted comparative experiments on robustness in low-expert-data regime in the obstacle avoidance task for the quadrupedal robot locomotion. The results, as shown in Table{~\ref{tab:appendix}}, show that our CasIL maintains the robustness of skill imitation even in a dynamic environment with low-expert-data.

\subsubsection{Effect of the Number of Options (Skills)}
\begin{figure}[htb!]
\centering
\includegraphics[width=0.8\columnwidth]{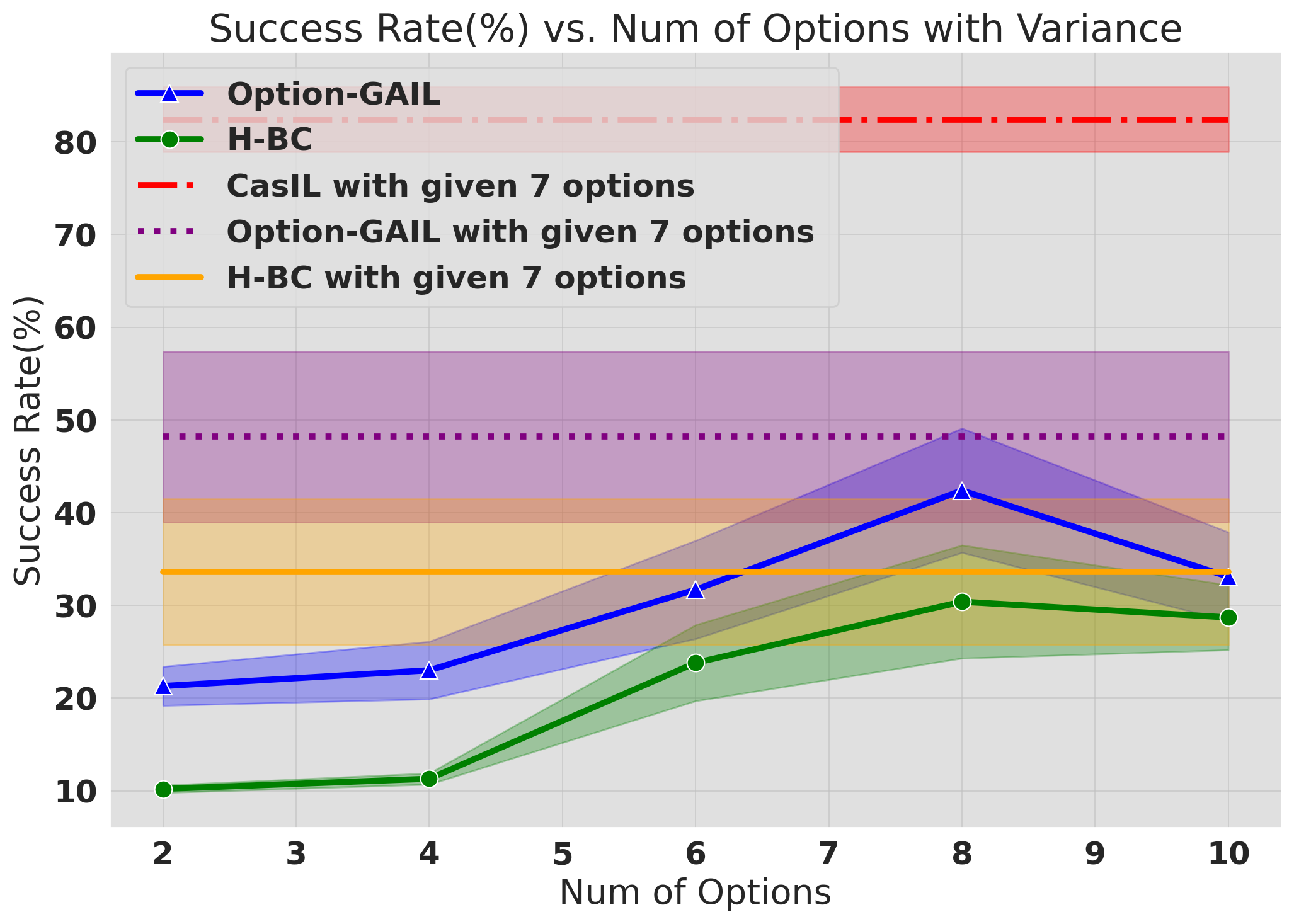} 
\caption{Effect of different number of options on the test success rate of Option-GAIL and H-BC in the \textit{PlayJenga} task.}
\label{options}
\end{figure}

Although in the Experiments Section we gave the same number of options as CasIL for Option-GAIL and H-BC for a fair comparison, we also wanted to know how the performance of Option-GAIL and H-BC would be affected if the number of options was different, and therefore we further evaluated the effect of a given number of options on Option-GAIL and HBC in the robotic arm manipulation task \textit{PlayJenga}. Fig.{~\ref{options}} illustrates the effect of the number of options on the test success performance of Option-GAIL and H-BC, where ``CasIL with given 7 options", ``Option-GAIL with given 7 options", and ``H-BC with given 7 options" are used as the comparison criterion. It can be clearly seen that either too few or too many options (skills) negatively affect the performance of skill imitation, whereas the optimal set of skills input through human cognitive priors leads to the relatively best skill imitation capbilities for the models.

\subsection{Preliminaries}
The mathematical basis of reinforcement learning (RL) is Markov Decision Processes (MDPs), which also applies to imitation learning (IL) algorithms for sequential decision-making tasks. Under the setting of RL, an MDP can be defined as a four-tuple: $(S,A,P,R)$, where $S$ represents a finite set of states, $A$ represents a finite set of actions, $P : S \times A \times S \to [0,1]$ represents the probability of the state transition, $P(s'|s,a)$ represents that an agent takes action $a$ at state $s$ then transfer to $s'$, and the reward function $R(s,a)$ represents the reward obtained by the agent after performing the action $a$ at state $s$. 
The goal of the agent in RL is to find a policy $\pi: S \times A \to [0,1]$ that can maximum the long-term cumulative reward. The discounted accumulate reward $J(\pi) = \mathbb{E}_\pi [\sum_{t=0}^T \gamma^t R(s_t,a_t)]$ is used to evaluate the quality of policy $\pi$, in which $a_t = \pi(s_t)$, $\gamma$ is the discount rate, $T$ is the limited time step, and the optimal policy is $\pi^* = \mathrm{argmax}_\pi J(\pi)$.
MDPs as traditionally conceptualized do not involve temporal abstraction or time-extended actions. They are based on discrete time steps: a single action taken at time $t$ affects state and reward at time $t + 1$. There is no notion of continuous action over a variable time period. As a result, traditional MDP approaches fail to take advantage of the simplicity and efficiency that higher levels of temporal abstraction can provide \cite{sutton1999between}.

One of the keys to using temporal abstraction as a minimal extension of the RL architecture is based on the theory of Semi-Markov Decision Processes (SMDPs). SMDPs are a special kind of MDPs suitable for modeling continuous-time discrete-event systems. Actions in SMDPs require variable time and are intended to simulate temporally extended processes of action.
In the high-level cognition generator for robotics applications in our main paper, the set of states is $\mathcal{S}$ and the set of actions is the set of options. Since MDPs are Markovian and options are semi-Markovian, these expectations and distributions are explicitly defined; thus, the next state, reward, and time depend only on the option and its state at the time of initiation. The duration of an option is always discrete and variable, which is one of the arbitrary temporally-extended intervals allowed in SMDPs.

Inspired by this, our work employs an option-based hierarchical structure and utilizes SMDP to model the decision-making process that encompasses skill cognition, thus enabling our entire architecture to perform cognition-action end-to-end IL training.

\begin{figure}[htb!]
\centering
\includegraphics[width=\columnwidth]{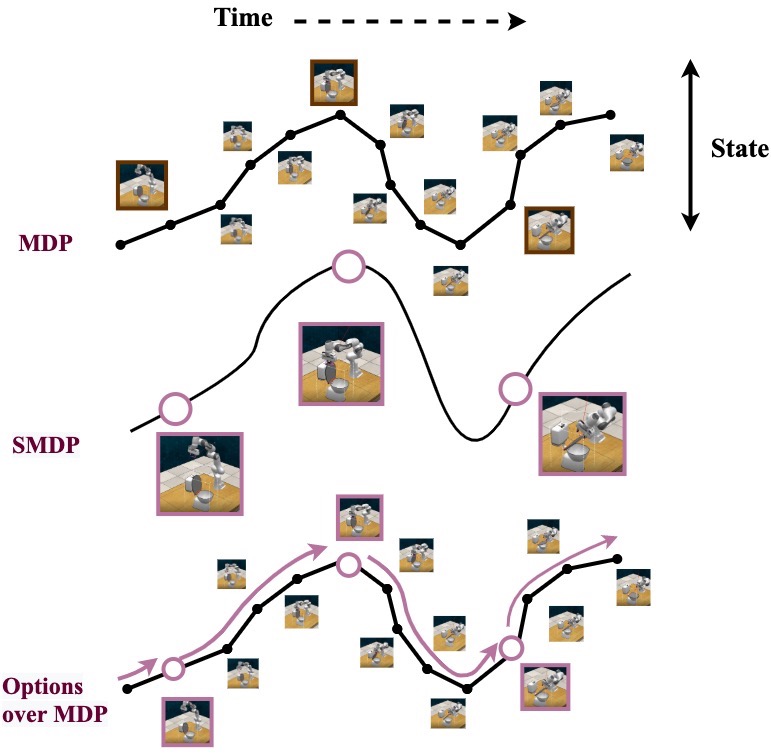} 
\caption{Using the \textit{Toiletseatdown} task in the main paper as an example, the state trajectory of the MDP consists of small discrete-time transitions, while the state trajectory of the SMDP consists of large continuous-time transitions. Options allow the trajectories of the MDP to be analyzed in both ways. The illustrative examples of MDP and SMDP in the figure refer to the form of Figure 1 in \cite{sutton1999between}.}
\label{smdp}
\end{figure}


\end{document}